\documentclass[a4paper]{article}
\usepackage{ISCSLP2026}
\usepackage{booktabs}
\usepackage{graphicx}
\usepackage{amsmath}
\usepackage{array}
\usepackage{newtxtext}
\usepackage{balance}
\usepackage[hidelinks]{hyperref}

\graphicspath{{figures/}}

\title{Patient-Specific Residual-Error Modeling in Personalized Mandarin Dysarthric ASR: Distributional Fidelity and Correction Diagnostics}
\name{Zhongzhen Ding \qquad Yuxuan Wu \qquad Zhaojie Luo\sthanks{\hspace*{0.6em}Corresponding author.}}
\address{Southeast University, China}
\email{213240580@seu.edu.cn, luozhaojie@seu.edu.cn}

\ifdefined
\fi

\begin{document}
\maketitle

\begin{abstract}
Personalized ASR can reduce errors in dysarthric speech, yet the substitutions, deletions, and insertions that remain after adaptation may still vary across patients. Recent studies have used synthetic data for ASR personalization, but mainly ask whether it improves recognition; whether it reproduces a target patient's post-adaptation error distribution remains unclear. We estimate patient-specific residual-error distributions from reference--hypothesis alignments and use them to generate corrupted text. With clean text, edit counts and locations, randomization, and generation budgets matched, we compare patient-specific and target-excluded population models using residual-error distribution fidelity---the agreement between generated and observed error distributions. Patient-specific generation better matched observed errors for seven of eight patients, showing that post-adaptation errors retain patient-specific structure useful for synthetic error design. In a matched correction diagnostic, training on patient-specific errors produced nearly five times as many exact repairs as training on population errors, but its higher edit rate introduced enough harmful changes to erase the net gain. These results support patient-specific residual-error modeling for synthetic text design while separating distributional fidelity from correction utility: net gains also require better correction candidates and selective edit acceptance.
\end{abstract}

\noindent\textbf{Index Terms}: dysarthric speech recognition, personalized ASR, residual errors, distributional fidelity, error correction

\section{Introduction}
Dysarthric speech varies substantially across patients, while labeled recordings from any one patient remain limited. Patient-specific adaptation can improve recognition from such data, but substitutions, deletions, and insertions remain after adaptation and may differ across patients \cite{Tobin2022Personalized,Jin2024Personalized,MullerEberstein2024Hypernetworks}. If these residual errors have repeatable patient-specific structure, additional data should reproduce the target patient's residual-error distribution, not merely increase the training set.

Synthetic data offers a controllable way to examine this requirement. Text selection affects TTS-based ASR augmentation, and in personalization content can matter more than matching the target voice \cite{Liu2023Selection,Yang2023Text}; dysarthric speech synthesis has supported few-shot and personalized training, but may complement rather than replace authentic recordings \cite{Hermann2023FewShot,Leung2024TTDS,Wagner2025Personalized,Li2025Supplement}. However, final ASR performance cannot isolate whether generated data reproduce a target patient's post-adaptation errors. This distinction matters for dysarthric ASR correction \cite{LaQuatra2025Correction}: mismatched training errors may not teach a model the errors it needs to handle \cite{Ghosh2025Failing}. Therefore, before discussing correction utility, we first ask whether the generated data faithfully represent the target patient.

We model patient-specific residual errors from alignments between reference transcripts and personalized ASR outputs, then use the estimated substitution, deletion, and insertion distributions to generate corrupted text. We define \emph{residual-error distribution fidelity} as how closely a generated distribution matches the patient's observed residual errors. The patient-specific and target-excluded population models share the clean text, edit count, edit location, randomization, and generation budget; only target-patient information differs.

With all other generation factors matched, patient-specific generation achieved higher distributional fidelity for seven of eight patients. A focused correction diagnostic showed why this did not yield a net gain: patient-specific modeling found more exact repairs, but extra edits introduced more errors than they removed. These findings separate two questions: whether synthetic errors represent the target patient and whether a correction system can use that representation safely.

\section{Method}
\begin{figure*}[!t]
  \centering
  \includegraphics[width=0.99\textwidth]{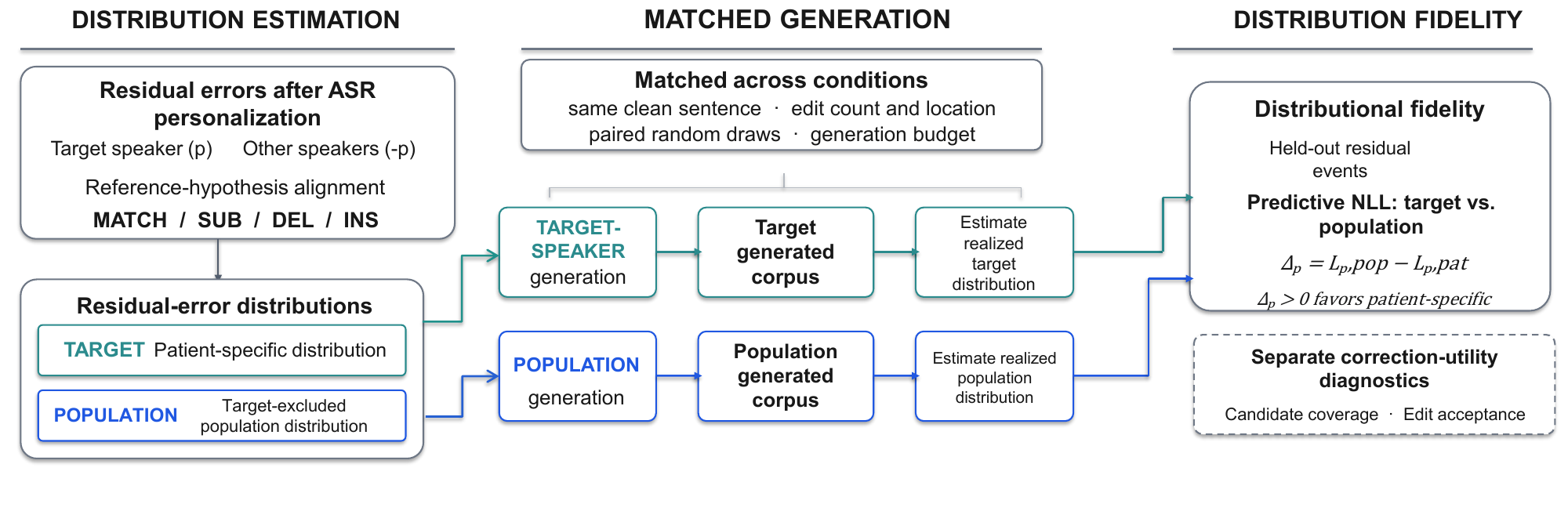}
  \caption{Three-stage method overview. (2.1) Reference--hypothesis alignments from the target and remaining speakers are used to estimate a patient-specific residual-error distribution and a target-excluded population distribution. (2.2) These distributions generate corrupted versions of identical clean sentences under matched edit counts and locations, paired random draws, and generation budgets. (2.3) The distributions realized by the generated corpora are compared using mean event-level predictive NLL on held-out target-speaker residual errors; positive $\Delta_p=L_{p,\mathrm{pop}}-L_{p,\mathrm{pat}}$ favors patient-specific generation. Correction utility is evaluated separately.}
  \label{fig:flow}
\end{figure*}

We test whether a synthetic error corpus preserves the residual-error pattern of a target speaker after personalization. The method follows the three linked stages shown in Figure~\ref{fig:flow}: Section~\ref{sec:dist-estimation} (distribution estimation) estimates the patient-specific and target-excluded population residual-error distributions from reference--hypothesis alignments; Section~\ref{sec:matched-generation} (matched generation) uses them to create paired corrupted texts while keeping the clean text, edit counts and locations, paired random draws, and generation budgets fixed; and Section~\ref{sec:fidelity-evaluation} (distribution fidelity) re-estimates the distributions realized by the generated corpora and compares their mean event-level predictive NLL on held-out target-speaker residual errors. Correction utility is evaluated separately because it additionally depends on candidate coverage and edit acceptance.

Figure~\ref{fig:flow} makes the controlled contrast explicit. In the primary comparison, the target-speaker and target-excluded population paths share the clean text, edit counts, edit locations, paired random draws, and generation budget; only the distribution used to generate errors changes. Because finite sampling and character realization can shift the realized corpus away from its sampling distribution, we re-estimate the distribution from each corpus before scoring it. The dashed branch denotes the separate correction diagnostic, whose outcome also depends on candidate coverage and edit acceptance.

\subsection{Residual-error distribution estimation}
\label{sec:dist-estimation}
For patient $p$, let $\mathcal D^{\mathrm{fit}}_p=\{(r_{pj},h_{pj})\}_j$ denote the reference--hypothesis pairs used to estimate residual errors, where $h_{pj}$ is produced by the patient-adapted ASR model on development utterances excluded from adaptation updates. We align $r_{pj}$ and $h_{pj}$ by minimum character edit distance, breaking ties in favor of more exact matches. The alignment yields exact matches, substitutions, deletions, and insertions. All four outcomes contribute to the edit-operation distribution, whereas fidelity is evaluated only on the three residual error types.

Exact character-to-character confusions are sparse for a single patient. Since phonological information is useful for Chinese ASR error correction \cite{Fang2022Phonological}, we represent each character by its Mandarin initial, final, and tone. The model first estimates the probabilities of an exact match, substitution, deletion, and insertion conditioned on the source features. For substitutions, it further estimates the conditional distributions of target initials, finals, and tones; for insertions, it estimates their marginal distributions.

Because each patient provides few residual errors, we regularize the patient estimate toward a population prior fitted after excluding that patient. Let $u$ denote the relevant conditioning context and $z$ an edit operation or target phonological feature. We estimate
\begin{equation}
  \widehat p_p(z\mid u)=
  \frac{n_p(z\mid u)+\kappa\widehat p_{-p}(z\mid u)}
       {N_p(u)+\kappa},
  \label{eq:smoothing}
\end{equation}
where $n_p(z\mid u)$ and $N_p(u)$ are target-patient counts and $\widehat p_{-p}$ is the target-excluded population distribution. This shrinkage retains patient-specific evidence when available and stabilizes estimates when it is sparse.

To avoid individualizing from very few errors, we perform leave-one-example-out prediction over the fitting pairs. Each pair is removed in turn, the patient model is re-estimated from the remaining pairs, and both models are scored on the omitted residual errors. The adaptive condition uses the patient estimate only when
\begin{equation}
  G_p=L^{\mathrm{LOO}}_{p,\mathrm{pop}}-L^{\mathrm{LOO}}_{p,\mathrm{pat}}>0,
  \qquad N^{\mathrm{err}}_p\geq5.
  \label{eq:gate}
\end{equation}
It otherwise uses the population estimate. Fixed patient and population conditions are retained for every patient to describe between-patient variation.

\subsection{Matched residual-error generation}
\label{sec:matched-generation}
The central comparison uses the patient-specific model and the target-excluded population model. The former includes the target patient's residual errors, whereas the latter uses only the remaining patients; their contrast therefore tests whether target-patient information changes the generated error distribution. Two auxiliary controls clarify this contrast. An unconditioned population baseline ignores source-character phonological features and uses overall error frequencies, while a mismatched-patient control uses another patient's error distribution. They test the roles of source conditioning and correct patient identity, respectively.

For each clean sentence, all generation conditions use the same number and locations of errors. At those locations, a model generates a substitution, deletion, or insertion. For substitutions and insertions, it then samples target initials, finals, and tones and selects a compatible character from an inventory derived from external text. The conditions also share the clean text, paired random draws, and number of generated examples, producing aligned clean--corrupted text pairs.

After generation, we realign each text pair and estimate a residual-error distribution from each realized corpus. Fixed edit counts and the realization of phonological features as specific characters can shift the final corpus away from the sampling parameters. The subsequent evaluation therefore uses the distribution induced by the generated corpus rather than the generator parameters alone.

\subsection{Residual-error distribution fidelity}
\label{sec:fidelity-evaluation}
We define \emph{residual-error distribution fidelity} as the predictive fit of a distribution estimated from generated text to the substitutions, deletions, and insertions actually observed after ASR personalization. Evaluation uses a separate set of residual errors. Lower mean event-level negative log-likelihood (NLL) indicates higher fidelity to the patient's errors.

For an observed residual error $e$, let $o_e$ be its edit operation, $u_e$ the corresponding source-phonological context, and $t_{e\lambda}$ the target feature at layer $\lambda\in\{I,F,T\}$, denoting initial, final, and tone. Insertions have no source character and therefore use the global edit-operation distribution. Let $\widetilde p_a$ denote the error distribution estimated from the corpus generated under condition $a$. The event loss is
\begin{equation}
\begin{aligned}
\ell_a(e)={}&-\log q\!\left(\widetilde p^{\mathrm{op}}_a(o_e\mid u_e)\right)\\
&-\mathbf{1}[o_e=\mathrm{SUB}]\sum_{\lambda\in\{I,F,T\}}\tfrac13
  \log q\!\left(\widetilde p^{\mathrm{sub}}_{a,\lambda}(t_{e\lambda}\mid u_{e\lambda})\right)\\
&-\mathbf{1}[o_e=\mathrm{INS}]\sum_{\lambda\in\{I,F,T\}}\tfrac13
  \log q\!\left(\widetilde p^{\mathrm{ins}}_{a,\lambda}(t_{e\lambda})\right),
\end{aligned}
\label{eq:eventloss}
\end{equation}
where $q(x)=\max(10^{-12},x)$. Deletions use only the operation term; substitutions and insertions also include three equally weighted target-feature terms. For patient $p$ with observed residual-error set $\mathcal E^{\mathrm{err}}_p$, we define
\begin{equation}
  L_{p,a}=\frac{1}{|\mathcal E^{\mathrm{err}}_p|}
  \sum_{e\in\mathcal E^{\mathrm{err}}_p}\ell_a(e),
  \qquad
  \Delta_p=L_{p,\mathrm{pop}}-L_{p,\mathrm{pat}}.
  \label{eq:estimand}
\end{equation}
Positive $\Delta_p$ indicates that patient-specific generation better matches the patient's observed residual errors. Fidelity asks whether generated text reproduces an error distribution; correction asks whether a model trained on generated errors can reduce errors without corrupting correct text. Because that outcome also depends on the correction model and its edit-acceptance rule, we evaluate it as a separate endpoint.

\section{Experimental setup}

\subsection{Speakers and residual-error data}
We analyzed eight speakers with dysarthria from the Chinese Dysarthria Speech Database (CDSD) \cite{Wang2024CDSD}. All personalized ASR models started from the same speaker-independent Paraformer-large model \cite{Gao2022Paraformer} and were fine-tuned using the same procedure and 100 real utterances per speaker. The base-model training data did not include any speakers from either study cohort. This common starting point and adaptation budget made residual errors comparable across speakers.

For each speaker, 51--161 reference--hypothesis pairs were used to estimate the residual-error distribution, and a disjoint set of 250--328 pairs was used for fidelity evaluation. Neither set contained the ASR adaptation utterances, and the two sets shared no utterances or exact reference--hypothesis pairs.

\subsection{Controlled text-error generation}
The four generation conditions shared source text, edit counts and positions, paired random draws, and generation budget. Only the probability distribution used to generate errors changed. Source text came from WenetSpeech4TTS Premium transcripts, not audio \cite{WenetSpeech4TTS2024}. After deduplication and removal of texts overlapping the CDSD material used here, we segmented the text with Jieba and extracted spans of one to four consecutive tokens. The spans were then matched by character length. Each speaker--condition pair comprised 6,000 examples in three non-overlapping batches.

The central comparison was between the target-speaker distribution and a population distribution estimated without that speaker. An unconditioned population control tested source-character phonological information. A mismatched-speaker control tested whether the correct speaker identity mattered.

We smoothed sparse probability estimates. For the target-speaker-excluded population and generated-corpus estimates, global probabilities used additive smoothing with $\alpha=0.5$, and source-conditioned probabilities were shrunk toward the corresponding global distribution with $\kappa=2$. Target-speaker estimates were shrunk toward the target-speaker-excluded population with $\kappa=50$. Initial, final, and tone received equal weight.

\subsection{Evaluation protocol and uncertainty}
Distributional fidelity was measured by mean event-level predictive negative log-likelihood (NLL) on observed residual errors. Lower NLL indicates a closer match.

The primary macro-average across speakers included only those with sufficient evidence for speaker-specific modeling. We left out one fitting pair at a time and compared how well the target-speaker and population models predicted its residual errors. A speaker was included when the target-speaker model performed better overall and at least five residual error events were available. We also report all eight speakers descriptively and use the two controls to test source-character information and speaker identity.

We obtained 95\% intervals by bootstrap resampling. Within-speaker intervals resampled paired evaluation examples. The primary macro interval first resampled speakers and then examples within each speaker. The descriptive eight-speaker interval resampled only fixed speaker-level contrasts. The three generation batches assessed stability, not additional speakers.

A separate six-speaker cohort was evaluated only if at least three speakers met the same prespecified applicability criterion.

{\raggedright
An anonymized code release is available at:\\[-1pt]
\url{https://github.com/deasy778/anonymous-residual-error-fidelity}\par}

\subsection{Correction-utility diagnostics}
Correction was evaluated as a separate endpoint with two complementary diagnostics based on the same publicly released BART4CSC checkpoint (revision 6bce2c5) \cite{Lewis2020BART,Xu2024BART4CSC}. For the first diagnostic, we fine-tuned two systems on matched synthetic corpora generated from either the target-speaker or target-excluded population error distribution. The systems shared their initialization, clean text, generation and training budgets, and evaluation procedure. On 2,000 corrupted D37 inputs and an equally sized unchanged set, an edit was accepted only when its confidence advantage over copying the input exceeded a threshold calibrated on other speakers.

The second diagnostic was a fixed-candidate upper-bound analysis on 810 reference--hypothesis pairs from the eight speakers. The public base model generated candidate corrections, with the original input retained as an option, and an oracle selected the candidate with the lowest reference error. This analysis measures the headroom within the available candidates rather than achievable correction performance.

\section{Results and Discussion}
\subsection{Target-speaker information better represents post-personalization residual errors}
\begin{figure}[!t]
  \centering
  \includegraphics[width=0.98\columnwidth]{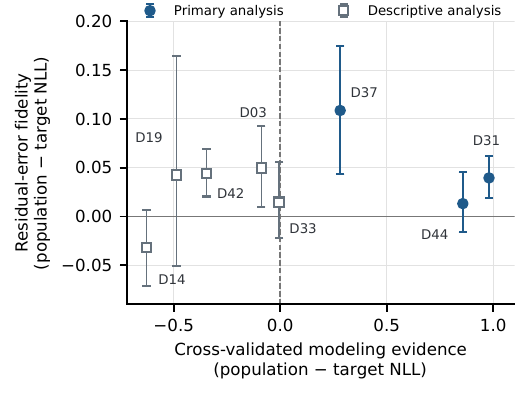}
  \caption{Speaker-level relation between cross-validated evidence for a target-speaker model (horizontal) and residual-error fidelity on a separate evaluation set (vertical). Positive fidelity contrasts favor target-speaker generation. Points show estimates; bars are paired-example 95\% intervals.}
  \label{fig:patient_fidelity}
\end{figure}

A controlled comparison yielded a clear positive result. With clean text, edit counts and locations, and paired random draws held fixed, errors generated from the target-speaker distribution more closely matched the observed residual errors for seven of eight speakers; speaker-macro predictive NLL decreased from 3.409 to 3.374. Although the gain varied across speakers, leaving out any one speaker preserved the improvement in the speaker-macro average. Residual errors therefore retained speaker-specific structure after ASR personalization that the target-excluded population distribution did not fully represent, and the pattern was not driven by a single speaker (Figure~\ref{fig:patient_fidelity}).

The advantage was further supported for the three speakers with stronger modeling evidence (Table~\ref{tab:main_source}). Target-speaker generation reduced their macro-average NLL from 3.017 to 2.963, with all three point estimates favoring the same direction. Matching deteriorated when source-character initials, finals, and tones were ignored or when another speaker's error distribution was used. The full-cohort pattern, focused analysis, and two matched controls jointly indicate that post-personalization residual errors are not merely generic ASR noise: source-character phonology and target-speaker identity both carry structure that improves synthetic-error fidelity. Useful synthetic data should therefore reproduce errors likely to be encountered by the personalized model, rather than merely increase error volume. This conclusion applies when the available residual errors support a speaker-specific distribution. Only one speaker in the independent cohort met the same minimum-evidence criterion, so no cohort-level fidelity comparison was performed.

\begin{table}[!t]
\centering
\caption{Residual-error fidelity in the primary three-speaker analysis. Lower NLL is better; positive differences favor target-speaker generation. The first four rows use the target-excluded population comparator; the final two use auxiliary controls.}
\label{tab:main_source}
\scriptsize
\setlength{\tabcolsep}{2.8pt}
\begin{tabular}{lrrrrl}
\toprule
Analysis & Events & Target & Comp. & Difference & 95\% interval \\
\midrule
D31 & 1,057 & 3.0130 & 3.0525 & 0.0395 & $[0.0187,\,0.0619]$ \\
D37 & 216 & 3.0290 & 3.1376 & 0.1086 & $[0.0434,\,0.1747]$ \\
D44 & 619 & 2.8465 & 2.8597 & 0.0131 & $[-0.0162,\,0.0455]$ \\
\midrule
Macro & --- & 2.9628 & 3.0166 & 0.0538 & $[0.0108,\,0.1132]$ \\
No phonology & --- & 2.9628 & 3.0746 & 0.1118 & --- \\
Wrong speaker & --- & 2.9628 & 3.0246 & 0.0618 & --- \\
\bottomrule
\end{tabular}
\end{table}

\subsection{Patient-conditioned synthetic errors expose more repair opportunities but require more selective correction}
This positive effect was not confined to the distribution metric. Starting from the same public Chinese correction model, we fine-tuned two systems on matched data generated from either the target-speaker or population error distribution. On 2,000 corrupted D37 texts, target-speaker-error training produced 48 exact repairs, nearly five times the 10 produced by population-error training. Patient-related synthetic errors therefore not only yielded a closer distributional match, but also triggered more genuinely correct corrections. At the same time, the target-speaker-trained system modified 617 inputs, compared with 117 for the population-trained system. This higher edit rate created more repair opportunities but also more harmful edits: added error characters increased from 19 to 125, and the total error count rose from 3,031 to 3,062 instead of falling to 3,021 as under population-error training (Table~\ref{tab:decision_diagnostics}). In practical terms, patient-related training made the system more likely to act on correctable text, but did not provide an equally reliable signal for when the input should be preserved.

The positive and negative findings are therefore two consequences of the same behavior: patient-related data expanded the set of repair opportunities acted upon, but the current decision rule applied this increased tendency to edit too broadly. A separate candidate-set upper-bound analysis found limited but nonzero correct repairs among the available alternatives. Net gains thus require both better candidate coverage and a decision rule that abandons the input only when evidence for a correction is sufficient. The present diagnostics cannot quantify the separate contributions of these two constraints, but they move the question from whether patient-related signal exists to how selectively it can be used.

\begin{table}[!t]
\centering
\caption{Correction diagnostics. The first row reports an eight-speaker fixed-candidate upper bound; the D37 rows compare matched fine-tuning on population- and target-speaker-generated errors. Exact repairs restore an input to its reference; added error characters are newly introduced errors.}
\label{tab:decision_diagnostics}
\scriptsize
\setlength{\tabcolsep}{2.0pt}
\begin{tabular}{lrrrr}
\toprule
Diagnostic & Texts & \shortstack{Error chars.\\in$\to$out} & \shortstack{Exact\\repairs} & \shortstack{Added error\\chars.} \\
\midrule
Candidate-set oracle (8 spk.) & 810 & $1{,}285\to1{,}237$ & 22 & --- \\
Population-generated (D37) & 2,000 & $3{,}031\to3{,}021$ & 10 & 19 \\
Target-speaker-generated (D37) & 2,000 & $3{,}031\to3{,}062$ & 48 & 125 \\
\bottomrule
\end{tabular}
\end{table}

\section{Conclusion}
We examined whether synthetic errors from a patient-specific model better represent post-personalization ASR errors than those from a population model fitted without the target speaker. Target-speaker distributions better matched observed errors for most speakers, and training on their synthetic errors yielded more exact repairs in a focused correction diagnostic. Residual errors therefore retain speaker-specific structure after personalization that can influence correction behavior. Yet this training also triggered more harmful edits, so extra repairs produced no net gain: representing patient errors and using them safely are distinct problems. Turning this fidelity advantage into net gains requires better correction candidates and sufficient evidence before accepting an edit.

\balance
\bibliographystyle{IEEEtran}
\bibliography{references}

\end{document}